\documentclass[lettersize,journal]{IEEEtran}
\usepackage{amsmath,amsfonts}
\usepackage{algorithmic}
\usepackage{algorithm}
\usepackage{array}
\usepackage[caption=false,font=normalsize,labelfont=sf,textfont=sf]{subfig}
\usepackage{textcomp}
\usepackage{stfloats}
\usepackage{url}
\usepackage{verbatim}
\usepackage{graphicx}
\usepackage{cite}
\usepackage[table,xcdraw]{xcolor}  
\usepackage{amssymb}
\usepackage{booktabs}
\begin{document}

\title{SPColor: Semantic Prior Guided Exemplar-based Image Colorization}

\author{Siqi Chen, Xueming Li, Xianlin Zhang, Mingdao Wang, Yu Zhang, Yue Zhang
\thanks{Siqi Chen, Mingdao Wang and Yu Zhang are with the School of Artificial Intelligence, Beijing University of Posts and Telecommunications, China (e-mail: sqchen@bupt.edu.cn;  wmingdao@bupt.edu.cn; zhangyu\_03@bupt.edu.cn)}
\thanks{Xueming Li, Xianlin Zhang and Yue Zhang are with the School of Digital Media and Design Arts, Beijing University of Posts and Telecommunications, China (e-mail: lixm@bupt.edu.cn; zxlin@bupt.edu.cn; zhangyuereal@163.com)}
}


\IEEEpubid{}

\maketitle

\begin{abstract}
Exemplar-based image colorization aims to colorize a target grayscale image based on a color reference image, and the key is to establish accurate pixel-level semantic correspondence between these two images. Previous methods directly search for correspondence over the entire reference image, and this type of global matching is prone to mismatch. Intuitively, a reasonable correspondence should be established between objects which are semantically similar. Motivated by this, we introduce the idea of semantic prior and propose SPColor, a semantic prior guided exemplar-based image colorization framework. Several novel components are systematically designed in SPColor, including a semantic prior guided correspondence network (SPC), a category reduction algorithm (CRA), and a similarity masked perceptual loss (SMP loss). Different from previous methods, SPColor establishes the correspondence between the pixels in the same semantic class locally. In this way, improper correspondence between different semantic classes is explicitly excluded, and the mismatch is obviously alleviated. In addition, SPColor supports region-level class assignment before SPC in the pipeline. With this feature, a category manipulation process (CMP) is proposed as an interactive interface to control colorization, which can also produce more varied colorization results and improve the flexibility of reference selection. Experiments demonstrate that our model outperforms recent state-of-the-art methods both quantitatively and qualitatively on public dataset. Our code is available at \url{https://github.com/viector/spcolor}.
\end{abstract}


\section{Introduction}
\IEEEPARstart{I}{mage} colorization aims to generate plausible colors for pixels in a grayscale image, and is useful in applications such as improving visual quality of grayscale photographs. As a well known ill-posed problem, the color for a grayscale object can be various while maintaining visual reality.

To deal with this ambiguous problem, the fully-automatic methods \cite{colorful,Chromagan,colorseedpoint,coltrans,WACV,scgan,10313314} directly map the grayscale image to color image based on its semantic features by learning large scale data. However, these methods ignore user's preference in colorization. To better consider the user's subjective feelings, the user-guided methods \cite{intercolor1,intercolor2,intercolor3,intercolor4,panetta2019fast} provide an interface for the user to define object's color by drawing color scribbles. And more recent deep-learning based methods \cite{zhang2017real,intercoloredge} combine user interaction with large database, but their effectiveness is still limited by additional human effort with aesthetic knowledge.

The exemplar-based methods colorize the target image under the guidance of a reference image which represents the user's preference. Some existing methods \cite{stylization,Color2Embed,hdrtransfer,semantic} mainly consider the global color distribution of the reference image, and leverage methods such as AdaIN (Adaptive Instance Normalization) \cite{adain} to colorize the target image in a style transfer approach. However, these approaches can easily mis-colorize the objects between the reference and target images. More exemplar-based methods are employed by building pixel-level semantic correspondence between reference and target images, and then transferring the color according to the correspondence \cite{tai2005local,ironi2005colorization,chia2011semantic,gupta2012image,variational,10182273}. However, this kind of methods are highly dependent on the quality of the reference image, whose lighting, viewpoint and content should be as similar as possible to the target image, while such an ideal reference is usually hard to find. 

\IEEEpubidadjcol

\begin{figure*}[!t]
  \centering
  \includegraphics[width=\linewidth]{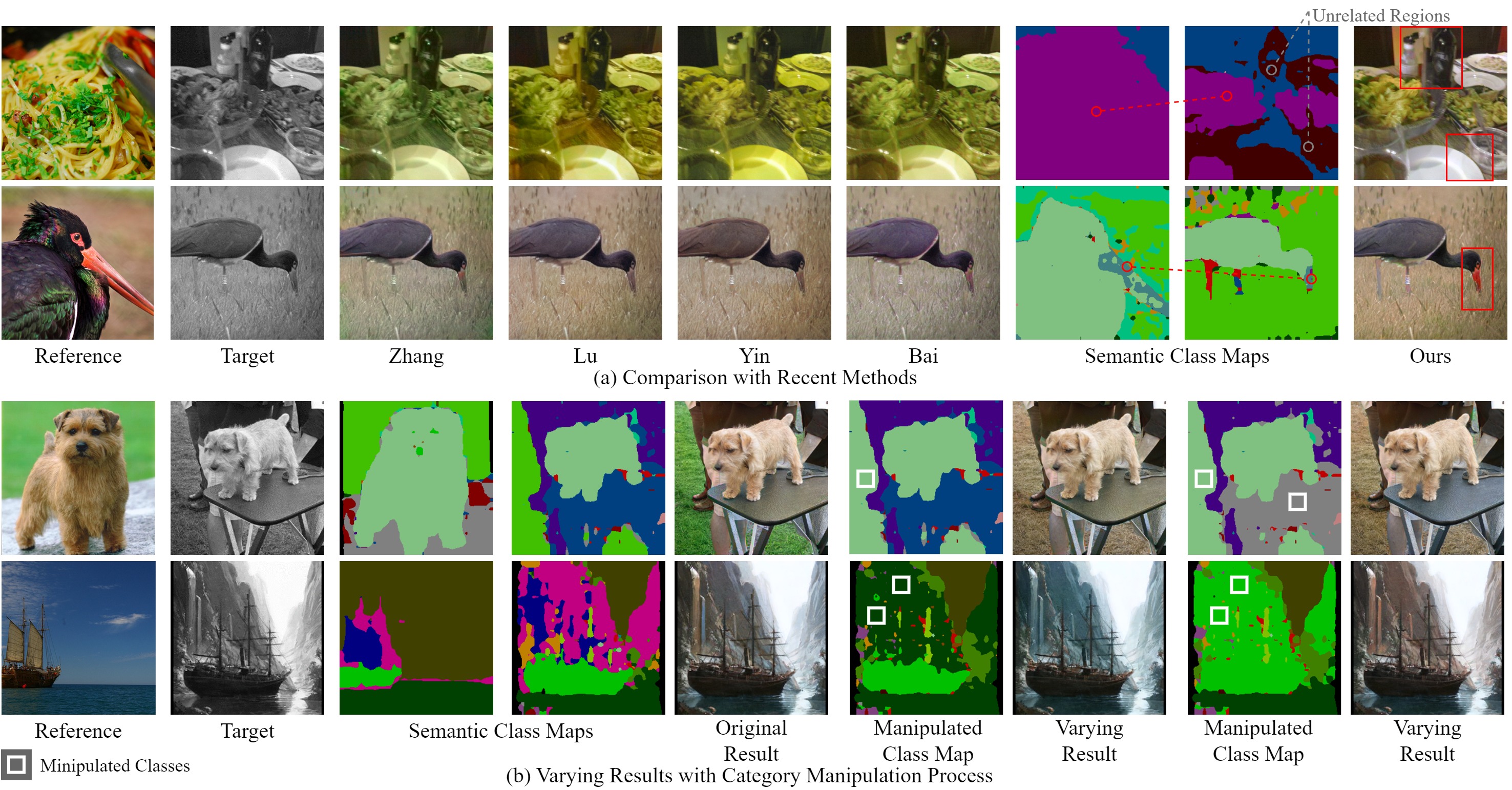}
  \caption{(a): Colorization results of recent methods of Zhang et al. \cite{DeepExamplar}, Lu et al. \cite{gray2colornet}, Blanch et al. \cite{xcnet}, Yin et al. \cite{transcolor}, and Bai et al. \cite{semantic} with ours. The red dashed lines connect the pixels which are corresponded, and the gray dashed lines denote the unrelated regions. Our method avoids the mismatch and produces more natural color.
  (b): Our varying results with category manipulation process. 
  }
  \label{fig:frontimg}
\end{figure*}

To overcome this issue, many approaches have been proposed \cite{DeepExamplarimage,DeepExamplar,gray2colornet,transcolor}. The basic assumption of these approaches is that the non-ideal reference mainly affects colorization in regions with low semantic correspondence confidence, so they utilize extra color databases or the global color distribution of the reference image to complete the color in these regions. However, in real scenarios, improper semantic correspondence also occurs in regions  with high semantic correspondence confidence, especially when with non-ideal reference. We summarize the reason for this type of mismatch in two aspects: (1) When the reference image contains only a part of objects related to the target image, improper correspondence will be established in unrelated regions and the color may be unnatural. As illustrated in the first column of {Fig. \ref{fig:frontimg}(a)}, recent methods make improper correspondence between food and tableware (which don't exist in the reference), and thus the tableware are colored by unnatural green or yellow;(2) It is prone to get mismatch in semantic correspondence when the scale, shape or texture of the objects are significantly different. As shown in the second column of {Fig. \ref{fig:frontimg}(a)}, the scale of the birds is very different, and it is challenging for recent methods to match the correct red color from the reference image.

Focusing on the mismatch problem in semantic correspondence, this paper proposes a novel exemplar-based image colorization framework named SPColor ({\bf S}emantic {\bf P}rior guided image {\bf Color}ization). Intuitively, a proper correspondence should be established between objects which are semantically similar. To achieve this goal, SPColor introduces the idea of combining semantic prior with colorization. It first leverages a newly designed category reduction algorithm and an off-shelf unsupervised semantic segmentation model \cite{stego} to classify input target and reference images to several pseudo-classes. Then the correspondence is conducted via the proposed semantic prior guided correspondence network (SPC). Unlike previous global matching, here the correspondence is established locally only between the pixels in the same class. In this way, the improper correspondence between different semantic classes is explicitly excluded, and the mismatch is obviously alleviated, as seen in {Fig. \ref{fig:frontimg}(a)}.  In addition, a novel similarity masked perceptual loss is designed to better reserve the color from the reference, which balances the color generation in regions with different correspondence similarities, and preserves the properly transferred color from the reference.  Furthermore, we propose a category manipulation process (CMP) as an interactive interface to control the class assignment in semantic segmentation, thus controlling the correspondence in SPC. By leveraging CMP, the correspondence can be established between pixels in each pair of classes according to the user's needs, which improves the flexibility of reference selection, and achieves varied results by a single reference image (Fig. \ref{fig:frontimg}(b)).

\begin{itemize}
  \item A novel framework named SPColor for exemplar-based image colorization is proposed. Utilizing semantic prior, SPColor transforms global matching mechanism into local matching in the core correspondence building step, thus alleviates mismatch evidently. Benefiting from the unsupervised semantic classification model, our method doesn't introduce additional training annotations compared with other state-of-the-art methods.
  \item SPColor is designed systematically. Several novel components are proposed to improve the color robustness, including a semantic prior guided correspondence network, a category reduction algorithm, and a similarity masked perceptual loss. 
  \item  We propose a category manipulation process as an interactive interface to control colorization according to the user’s need. It improves the flexibility of reference selection, and enables SPColor to achieve varied results by a single reference image.
  \item Experiments demonstrate that our model outperforms recent state-of-the-art methods both quantitatively and qualitatively on public dataset. Particularly, it is more robust and can generate more natural colors for non-ideal reference image. 
\end{itemize}

\section{Related Works}

In this section, we introduce the two main methods in exemplar-based image colorization: style transfer based methods and semantic correspondence based methods.
\subsection{Style transfer based methods}
For exemplar-based image colorization, some existing methods \cite{stylization,Color2Embed,hdrtransfer,semantic} mainly consider the global color distribution of reference image, and leverage methods such as AdaIN \cite{adain} to colorize the target image in a style transfer approach.

Xu et al. \cite{stylization} first use AdaIN for feature matching and blending in exemplar-based colorization. They propose a two-subnet architecture that achieves faithful colorization with a related reference image, and can also predict plausible colors with an unrelated one. 
In \cite{Color2Embed}, thin plate splines (TPS) \cite{duchon1977splines,chui2000new} transformation is adopted to create self-reference images, and the training approach are then implemented in a paired supervised manner. Besides, a progressive feature formalisation block is also designed for color injection. To handle HDR images, \cite{hdrtransfer} introduce a novel GAN-based method and propose to train networks in a self-supervised way. And for making the model focus on the semantically important regions in reference, \cite{semantic} proposes a sparse attention mechanism, and combines global color transfer (AdaIN based) with local details transfer (semantic correspondence based) for better performance. 
By leveraging methods like AdaIN, these methods can achieve fast processing speed. However, they mainly focus on the color style transfer, and are usually easy to mis-colorize the objects between reference and target images, which may violate the user's preferences.

\subsection{Semantic correspondence based methods}
Compared with AdaIN based methods, more exemplar-based methods are employed by building pixel-level semantic correspondence between reference and target, then transfer the color according to the correspondence \cite{tai2005local,ironi2005colorization,chia2011semantic,gupta2012image,variational}. However, in these methods, to make proper correspondence, the reference image should be as similar to the target image as possible, while an ideal reference is usually hard to find. 

In \cite{DeepExamplarimage}, the color database is first used to complete the color of unrelated regions (where the semantic correspondence confidence is relatively lower) by learning from large scale training data. And \cite{DeepExamplar} further enhance the performance by joint training the correspondence subnet and colorization subnet. In \cite{gray2colornet}, to make the result more similar to the reference, the color statistical information of the reference is used to predict the color in unrelated regions. And \cite{transcolor} further combines the semantic features in these regions, and proposes an attention based framework to unify the color prediction from semantic correspondence, the global color distribution of reference and the color database. Though remarkable achievement have been made, they usually ignore the mismatch problem in semantic correspondence and generate unnatural colors. It is worth mentioning that the recent works \cite{updown,semantic} also work toward preventing mismatch problem. For \cite{updown}, we claim that we are different in 3 aspects: (1) \cite{updown} learns the vertical semantic relation in reference and target images and works only for up-down mismatch problem, while our approach is spatially arbitrary, and can also deal with the issue that the reference contains only a part of related objects. (2) \cite{updown} generate semantic clusters via k-means according to the ab channel in original image pixels, while we generate clusters via deep neural networks. (3) \cite{updown} corrects the mismatch after correspondence by learning reference image's spatial distribution, while our approach prevents mismatch's happening before correspondence. And \cite{semantic} proposes a Semantic-Sparse Correspondence (SSC) module, which mainly uses the class activation map (CAM) of the reference image to reduce the correspondence interference caused by semantically insignificant areas. However, SSC still forces the pixels in target image to be corresponded with the reference image, even it is improper. Besides, though SSC observably avoids the mismatch for the main object by the CAM, the mismatch of other objects is reserved. Thus, how to obtain more accurate and robust correspondence for image colorization is still a challenging task.

\begin{figure*}[!t]
  \centering
  
  \includegraphics[width=\linewidth]{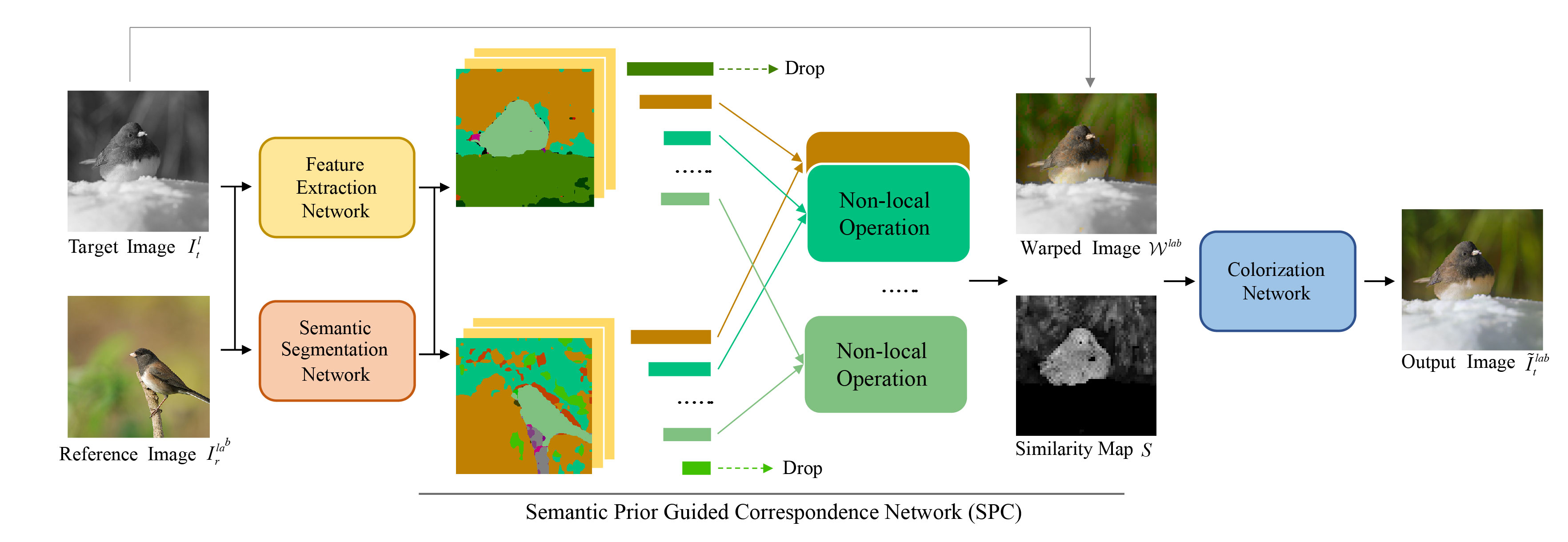}
  
  \caption{Overall framework of our approach. Firstly, both the reference and target images are sent to a feature extraction network and a semantic segmentation network, obtaining the features and pixel-wise classification labels. According to the classification labels (pseudo-label), the extracted features are then divided into several feature vectors. And in our semantic prior guided correspondence network (SPC), the feature vectors which belong to the same class will go through the non-local operation to find its semantic correspondence. In this way, we obtain the warped color image and the similarity map. Finally, the colorization network fills out the colors in unrelated regions, corrects the bad warped results and enhance the color naturalness.}
  \label{spcolor}
\end{figure*}

\section{Method}
\subsection{Problem formulation}
Given a grayscale target image $T^l$ and a reference image $R^{lab}$, exemplar-based image colorization aims to transfer the colors from reference image to the corresponding regions in target image, and generate a colorized target image $\hat T^{lab}$, where $l$ and $ab$ represents the luminance and chrominance channel in CIELAB color space respectively. To this end, the semantic correspondence is usually computed by the features of reference and target images extracted from a pretrained image classification network \cite{DeepExamplarimage,DeepExamplar,gray2colornet,transcolor} (e.g. VGG-19 network \cite{vgg}). The feature maps of reference and target images can be denoted as $F_T$ and $F_R$ respectively. And the dense correspondence $\mathcal{M}$ can be then computed by pairwise cosine similarity, which can be formulated in Eq. (1):
\begin{equation}
  \begin{aligned}
    \label{deqn_ex1}
    \mathcal{M}(i, j) = \frac{F_T(i)\cdot F_R(j)}{\|F_T(i)\|_{2}\|F_R(j)\|_{2}}
  \end{aligned}
\end{equation}
where $i$,$j$ represents the $i$-th and $j$-th position in target and reference feature maps respectively. The higher the $\mathcal{M}(i, j)$, the more similar of the corresponding pixels. For selecting the most similar position in reference for each of the position in target, each row of $\mathcal{M}$ is transformed to a one-hot vector via $\operatorname{Softmax}$ operation, and the most similar positions are then found by:
\begin{equation}
  \label{deqn_ex1}
  \mathcal{W}^{a b}(i)=\sum_{j} \operatorname{Softmax}(\mathcal{M}(i, j) / \tau) \cdot R^{a b}(j)
\end{equation}
In which $\tau$ represents a hyperparameter far less than 1, and $\mathcal{W}$ is the warped color map. And the similarity map value $S_w$ represents correspondence confidence can be easily defined by the maximal value in each line of $\mathcal{M}$:
\begin{equation}
  \label{deqn_ex1}
  \mathcal{S}_w(i)=\max_{j} \mathcal{M}(i, j)
\end{equation}
However, the warped color map $\mathcal{W}^{a b}$ may not be accurate in every position. Thus, a colorization network $\mathcal{F}_{color}$ is then responsible for correcting the bad warped results and improving the naturalness of the image with respect to the similarity map value $S_w$, as formally described in { Eq. (4)}:
\begin{equation}
  \label{deqn_ex1}
  \hat T^{lab} = \mathcal{F}_{color}(\mathcal{W}^{a b},S_w,T^l)
\end{equation}
The above  Eq. (1-3) compose to the vanilla non-local operation \cite{non-local}, which is widely used in colorization tasks \cite{DeepExamplarimage,DeepExamplar,gray2colornet,transcolor}. Though performs well in most cases, it is not robust enough under some circumstances. Firstly, as formulated in {Eq. (1-3)}, every pixel in target will be corresponded to a pixel in reference, even though they are semantically unrelated. And the similarity map may not be robust enough to help to fix the unnatural color in these regions.
Besides, since the correspondence relies on the features extracted from a pretrained encoder, we empirically find that it is easy to get mismatch especially when the scale, shape or texture of the objects
are significantly different. Different from the first situation, there are exact semantically related pixels in the reference image, but the correspondence still gets mismatch.

\begin{figure}[!t]
  \centering
  \includegraphics[width=\linewidth]{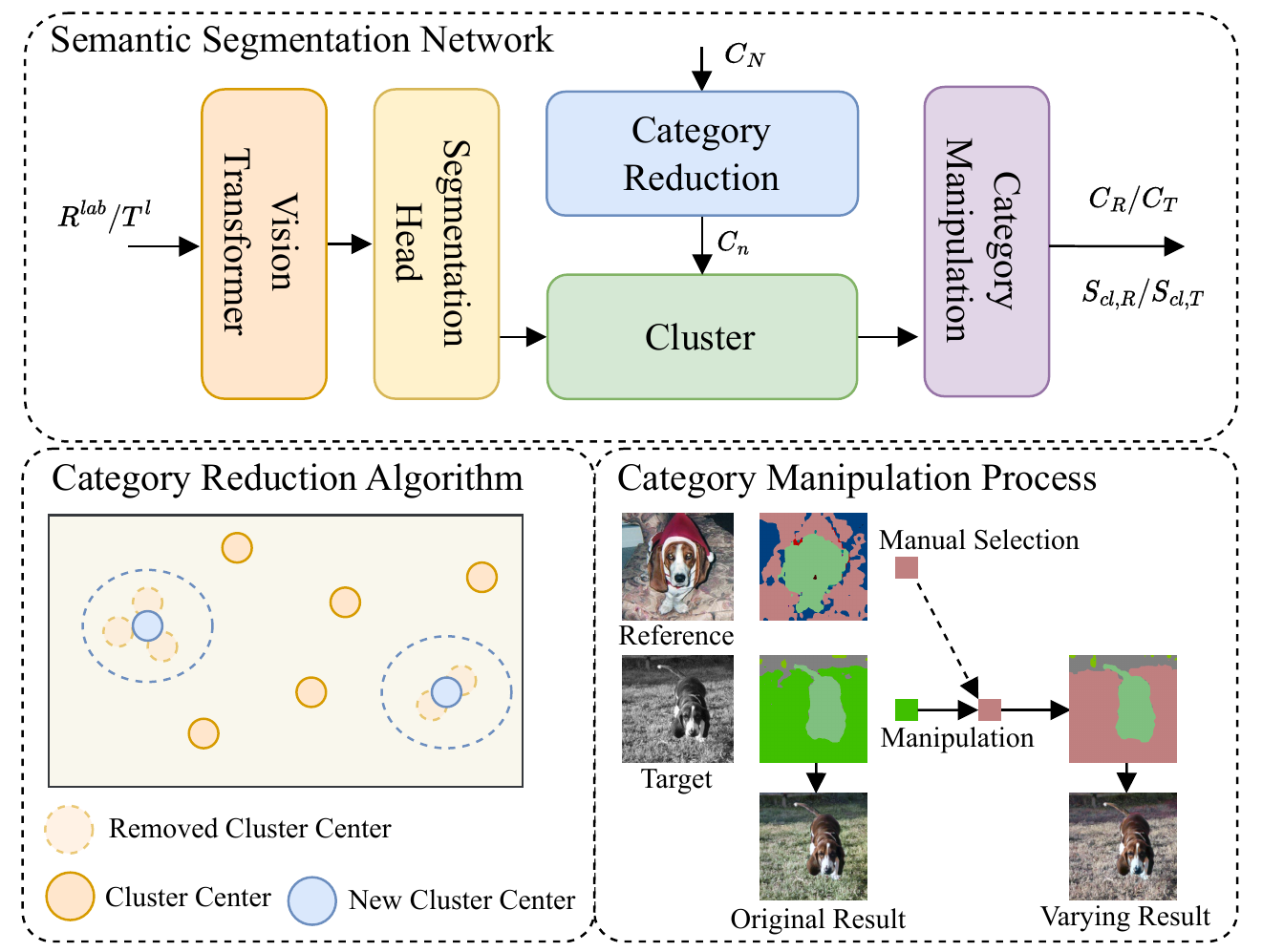}
  \caption{Structure of semantic segmentation network together with the illustration of category reduction algorithm and category reduction process. $C_N$ denotes the original set of pseudo-classes, which is then reduced to $C_n$ by our category reduction algorithm.}
  \label{fig:segnet}
\end{figure}

To overcome these issues, we propose a novel colorization framework with a semantic prior guided correspondence network (SPC), which is illustrated in {Fig. \ref{spcolor}}. Firstly, both the reference and target images are sent to a feature extraction network and a semantic segmentation network (with CRA and optionally CMP, as illustrated in Fig. \ref{fig:segnet}) to obtain the features and pixel-wise classification labels. According to the classification labels (pseudo-label), the extracted features are then divided into several feature vectors. And in our SPC, the feature vectors which belong to the same semantic class will go through the non-local operation to find its semantic correspondence. In this way, we obtain the warped color image and the similarity map. Finally, the colorization network fills out the colors in unrelated regions, corrects the bad warped results and enhance the color naturalness.

\subsection{Semantic prior guided correspondence network}

In order to resolve the two problems mentioned before, we propose a novel semantic prior guided correspondence network (SPC). 

Together with the feature extraction, a pretrained semantic segmentation network is used to cluster the pixels in reference and target images into several classes according to their semantic meanings. We denote the semantic class maps as $C_R$ and $C_T$ for reference and target images. And each pixel in the semantic class map is classified into $C_n \in \{C_0,C_1,...,C_k\}$, where $k$ is the number of classes. Besides, we get the classification confidence map $S_{cl,T}$ and $S_{cl,R}$, whose value measures the confidence for a pixel's classification.

It is worth mentioning that we don't care about the actual class label for a pixel, but only care about whether pixels belong to the same semantic class. Thus, the unsupervised semantic segmentation network is adopted, which doesn't introduce additional training annotations. According to the semantic class label (pseudo-label), the pixels in feature map $F_T$ can be segmented into several subsets via their classes.
\begin{equation}
  \label{deqn_ex1}
  F_{T,C_n} = \{F_T(i)|C_T(i) \in C_n\}, \quad n \in \{0,1,...,k\}
\end{equation}
where $F_{T,C_n}$ represents the pixels of $F_T$ in class $C_n$, and $i$ is the position in $F_T$. $F_{R,C_n}$ can be obtained in the same way. After classifying the pixels in feature maps into multiple classes, we can perform  class independent correspondence. The pixel regions of reference and target images in the same pseudo-classes are first paired, and then local correspondence within the class is performed. Thus, the semantic different pixels which have similar low-level features will not be considered. One may notice that, there may have mutually exclusive classes between $F_{T,C_n}$ and $F_{R,C_n}$. In this case, we simply drop the these classes and only calculate similarities in intersecting classes $C_{inter} =\{C_n|F_{T,C_n}\!\neq\!\varnothing, F_{R,C_n}\!\neq\!\varnothing\}$. Then the { Eq. 1} can be reformulated as:
\begin{equation}
  \begin{aligned}
    \label{deqn_ex1}
    \mathcal{M}_{C_n}(i, j) = \frac{F_{T,C_n}(i)\cdot F_{R,C_n}(j)}{\|F_{T,C_n}(i)\|_{2}\|F_{R,C_n}(j)\|_{2}}, \quad C_n \in C_{inter}
  \end{aligned}
\end{equation}
For the regions in target image which don't belong to the intersecting classes, we define them as \emph{unrelated regions}, where the warped color map $\mathcal{W}^{ab}$ and similarity map $S_w$ are both set to 0. In this way, the unrelated regions are explicitly defined. For the $i$-th position in $\mathcal{W}^{ab}$. We reformulated the { Eq. (2)} as:
\begin{equation}
  \label{deqn_ex1}
  \left\{
      \begin{array}{lr}
      \mathcal{W}^{a b}(i)= \mathcal{F}_{warp}(R_{C_T(i)}^{a b}), & C_T(i) \in C_{inter} \\
      \mathcal{W}^{a b}(i)= 0, & C_T(i) \notin C_{inter} 
      \end{array}
  \right.
\end{equation}
\begin{equation}
  \label{deqn_ex1}
  \mathcal{F}_{warp}(X)=\sum_{j} \operatorname{Softmax}(\mathcal{M}_{C_T(i)}(i, j) / \tau) \cdot X(j)
\end{equation}
\begin{equation}
  \label{deqn_ex1}
  R_{C_n}^{ab}(j) = \{R^{ab}(j)|C_R(j) \in C_n\}, \quad n \in \{0,1,...,k\}
\end{equation}
$R_{C_n}^{ab}$ is a subset of $R^{ab}$ where the pixels are in class $C_n$. And the { Eq. (3)} can also be rewritten.
\begin{equation}
  \label{deqn_ex1}
  \left\{
      \begin{array}{lr}
      \mathcal{S}_w(i)=\max_{j} \mathcal{M}_{C_T(i)}(i, j), & C_T(i) \in C_{inter} \\
      \mathcal{S}_w(i)=0, & C_T(i) \notin C_{inter}
      \end{array}
  \right.
\end{equation}

\subsection{Category reduction algorithm}
With SPC, the network has been able to more accurately match reference and target images based on the semantic prior. However, a natural concern is how finely we need to divide the categories of images (i.g. what is the number of $k$). On the one hand, it is necessary to have enough categories for SPC to avoid mismatch. On the other hand, as the number of categories grows, the probability of misclassification also increases. For unsupervised semantic segmentation models \cite{picie,stego}, they usually train a set of cluster centers to assign the extracted features to mid-level pseudo-classes (e.g. 27 classes in the CocoStuff \cite{cocostuff} and Cityscapes \cite{cityscapes} datasets). In our task, since we don't need to classify the images over fine grained, we start the class number at 27 following these methods, and further perform a category reduction algorithm (CRA) to adjust the class assignment. Specifically, we conduct an additional K-Means clustering on the trained cluster centers and reduce the class number to $k$ before clustering the extracted features (Fig. \ref{fig:cra}). The cluster centers that are close to each other are grouped into new ones, thus removing the easily confused classes for SPC. In this way,  the number of classes becomes a controllable coefficient in each inference step.

\subsection{Category manipulation process}
Our correspondence between the reference and target images are constrained by the semantic prior. Such method prevents the mismatch between objects in different semantics, and improves the visual quality of results. However, along with the more accurate matching, it also raises the difficulty of finding a proper reference image, which should be more similar to the target image. To solve the inflexible of SPC, we further design a category manipulation process (CMP). For each of the semantic classes in reference and target images, the user can change it to a manually selected class $C_p$ after clustering, and before SPC.
\begin{equation}
  \label{deqn_ex1}
  C_X(i) = C_p, \quad p\in \{0,1,...,k\} 
\end{equation}
where $X\in\{R,T\}$. The purpose is to retain the necessary constraints while having the flexibility to manipulate the constraints that do not degrade image quality. Fig. \ref{fig:segnet} shows an example of CMP. In the reference and target images are puppies on blanket and grassland respectively. Originally, the correspondence will only be established between the puppies, and the blanket will not transfer its color to grassland as they are belong to different semantic classes. However, one may believe that the brown color from the blanket is also considerable for the grassland. With the CMP, the semantic class of grassland in target image can be manipulated to the class which the blanket belongs to. Thus, the correspondence will be successfully established.
\subsection{Loss}
Our network aims to colorize grayscale target image via a colored reference image. The colorization result is not only supposed to be perceptually natural, but also should be similar to the reference image. To this end, several existing loss functions are adopted in our method, and a novel similarity masked perceptual (SMP) loss is designed.

{\bf SMP loss}. Perceptual loss measures semantic distance by comparing high-level features extracted in pretrained VGG-19\cite{vgg} network, and has been proved to be robust to appearance differences caused by two
plausible colors \cite{DeepExamplarimage}. We also empirically find that perceptual loss is effective in generating plausible colors in unrelated regions. Since the ground truth image corresponding to a certain pair of reference and target images does not exist, we can only do calculation between image result $\hat T$ and the original color image $T$, in which the color is unrelated to the reference image.  
\begin{equation}
  \label{deqn_ex1}
  \mathcal L_{perc} = \| \Phi_L(\hat T) - \Phi_L(T) \|_2^2
\end{equation}
where $\Phi_L$ represents activation map extracted at $L$-th layer. We set $ L = relu5\_2$ since the top layers contain more semantic information. Though robust to the appearance differences, the perceptual loss still lead the image result to be dissimilar to the reference. In this paper, we propose a simple improvement towards perceptual loss to enhance the reference-target similarity.

As described in the last subsection, we have similarity map $S_w$ measuring correspondence confidence, and $S_{cl,T}$, $S_{cl,R}$ measuring classification confidence. Firstly, for the reference image's classification confidence map $S_{cl,R}$, we align its value according to the semantic correspondence. For the $i$-th pixel in the aligned confidence map $S'_{cl,R}$:

\begin{equation}
  \label{deqn_ex1}
  \left\{
      \begin{array}{lr}
      S'_{cl,R}(i)= \mathcal{F}_{warp}(S_{cl,R,C_T(i)}), & C_T(i) \in C_{inter} \\
      S'_{cl,R}(i)= 0, & C_T(i) \notin C_{inter} 
      \end{array}
  \right.
\end{equation}

\begin{equation}
  \label{deqn_ex1}
  S_{cl,R,C_n}(j) = \{S_{cl,R}(j)|C_R(j) \in C_n\}, \quad n \in \{0,1,...,k\}
\end{equation}
Then, the final confidence map $S$ can be obtained by elementary multiplication of the three confidence maps:
\begin{equation}
  \label{deqn_ex1}
  S = S_w \odot S_{cl,T} \odot S'_{cl,R}
\end{equation}
$S$ is regarded as the final similarity map sent to the colorization network. The higher the value of $S$, the more reliable correspondence is established, where the reference's color should be preserved. Conversely, the correspondence is more unreliable, where the generation should be implemented. Thus, the SMP loss is designed as:
\begin{equation}
  \label{deqn_ex1}
  \mathcal L_{SMP} = (1-S) \odot \mathcal L_{perc} 
\end{equation}
In this way, the SMP loss focus more on regions with low correspondence confidence, especially the unrelated regions where $S$ equals to 0. And it gives less punishment on high confidence regions, which preserves the reference's color as far as possible.

{\bf L1 loss}. In training, we randomly provide the original colorized target image $T^{lab}$ as the reference image. In this case, we can use $T^{lab}$ as the color ground Truth, and we empirically find that this will make the image result more colorful. In order to determine pixel level colorization accuracy, we adopt L1 loss which computes pixel difference between chrominance of $\hat T^{ab}$ and the ground truth $T^{ab}$. We use L1 loss since it is demonstrated to generate more distinct color than L2 loss\cite{long2016learning,mathieu2015deep,niklaus2017video}. The L1 loss can be written as:
\begin{equation}
  \label{deqn_ex1}
  \mathcal L_{L1} = \|   \hat T^{ab} - T^{ab} \|_1
\end{equation}

In addition, we also adopt the adversarial loss  and smooth loss introduced in \cite{DeepExamplar} (using a single image instead of a video sequence as input), which help to generate vivid image and penalize color bleeding. Finally, our total objective loss can be written as:
\begin{equation}
  \begin{aligned}
  \mathcal L_{total}  = & \lambda_{L1} \mathcal L_{L1} + \lambda_{SMP} \mathcal L_{SMP}  + \lambda_{adv} \mathcal L_{adv} \\
  & + \lambda_{smooth} \mathcal L_{smooth}
 \end{aligned}
 \label{deqn_ex1}
\end{equation}

where the $\mathcal{L}_{adv}$, $\mathcal{L}_{smooth}$ represent the adversarial loss and smooth loss. And $\lambda_{L1}$, $\lambda_{SMP}$, $\lambda_{adv}$, and $\lambda_{smooth}$ represent the corresponding loss weights.

  
  

\section{Implementation} 

{\bf Datasets.} The datasets used in our experiment follow the settings of \cite{DeepExamplarimage,DeepExamplar,gray2colornet,Color2Embed,transcolor}. In training, we utilize the entire ImageNet ILSVRC 2012 dataset \cite{imagenet}, which involves more than 1.2 million of images from 1000 categories. Since there are grayscale, low-resolution or other improper images in part of the dataset, a data preprocessing is implemented. And finally, we get 1.02 million of training images. The reference images are selected via a searching algorithm proposed in \cite{DeepExamplarimage}. 

And in testing, we use the ImageNet 10k \cite{10k} dataset, which is a subset of the ImageNet validation set, containing 10 thousand of images. It excludes any grayscale single-channel images. 

{\bf Network structure.} Our feature extraction network is in the VGG-19 structure. And the colorization network is an encoder-decoder structure with skip connections, group convolutions, and dilated convolutions~\cite{dilatedconv}. With the remarkable progress in unsupervised semantic segmentation task \cite{picie,stego}, both the grayscale and the colorized images can be well segmented in unsupervised way. In this paper, we use \cite{stego} as our pretrained semantic segmentation network, and the images are segmented to 22 classes ($k=22$). We also tried supervised networks, but it is found to make few differences to the colorization results. As mentioned before, our method may not care of the actual  classification labels, but concerns which pixels are in the same class. Please refer to our published codes for more implementation details.

{\bf Training.} We randomly use (with a 20\% probability) the original colorized target image as the reference, which is empirically found useful to generate colorful result. And in this case, the hyperparameter $\lambda_{L1}=2.0$, otherwise $\lambda_{L1}=0$. Both the reference and target images are rescaled to $256 \times 256$ before sending to the networks.

The other hyperparameters are set as: $\lambda_{perc}=0.01$, $\lambda_{adv}=0.4$ and $\lambda_{smooth}=2.0$. We train the network for 120k iterations using the Adam optimizer with parameters $\beta_1=0.5$, $\beta_2=0.999$. The learning rate is set to $2 \times 10^{-4}$, $2 \times 10^{-5}$ for the discriminator and others, decayed by 10 at 100k iteration.

\begin{figure*}[t!]
  \centering
  \includegraphics[width=\linewidth]{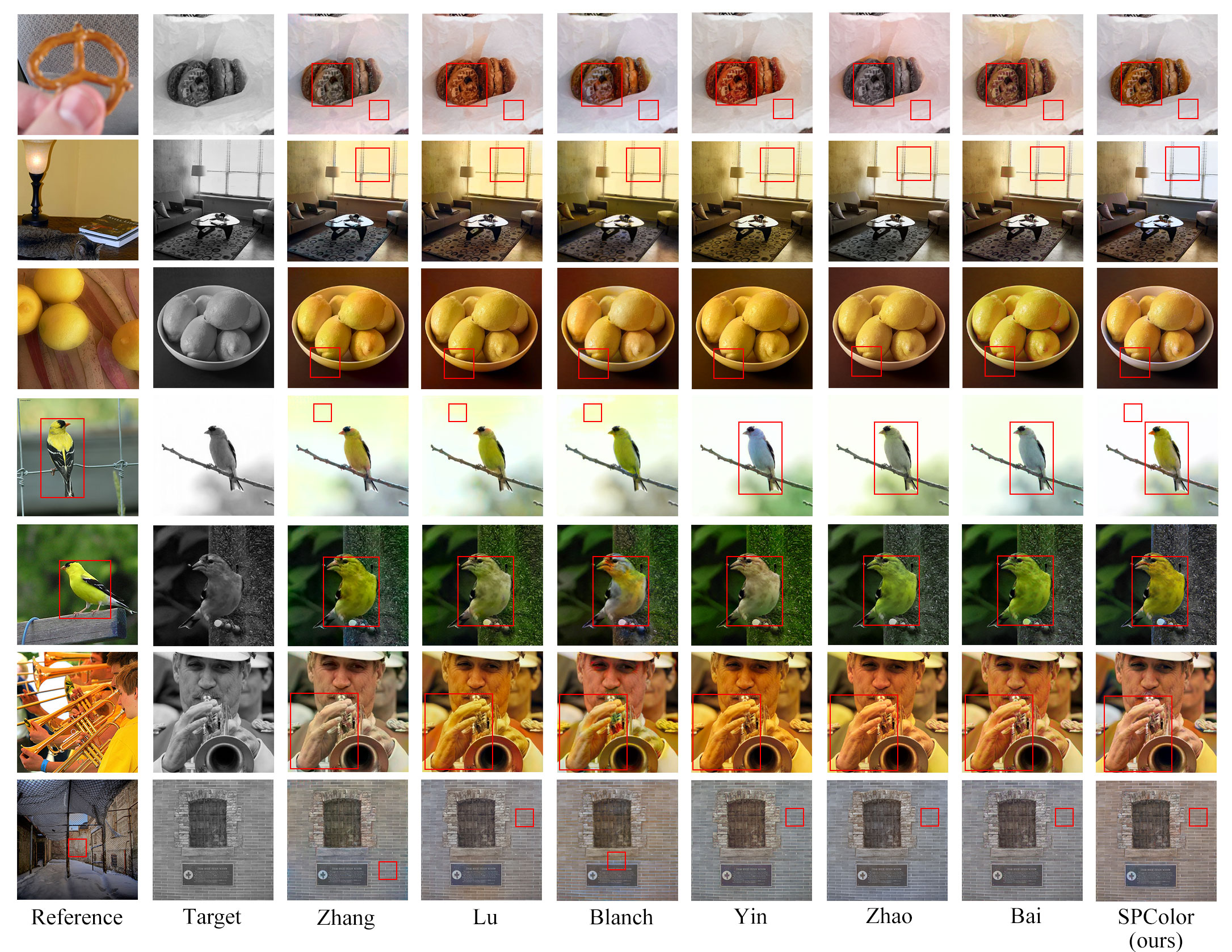} 
  
  \caption{More visual comparison with the state-of-the-art methods on ImageNet 10k dataset. The first column and the second column are the reference and target images. And from the third column to the eighth column are the colorization result of Zhang et al. \cite{DeepExamplar}, Lu et al. \cite{gray2colornet}, Blanch et al. \cite{xcnet}, Yin et al. \cite{transcolor}, Zhao et al. \cite{Color2Embed}, Bai et al. \cite{semantic}, and our SPColor respectively. }
  \label{appd:compare}
\end{figure*}

\begin{table}[]
  \centering
  \caption{Quantitative comparison with state-of-the-art exemplar-based colorization methods on ImageNet 10k dataset. Our method gives the best results on the all three metrics of image quality.}
  \label{tab:comparision}
  \begin{tabular}{
  >{\columncolor[HTML]{FFFFFF}}l |
  >{\columncolor[HTML]{FFFFFF}}c |
  >{\columncolor[HTML]{FFFFFF}}c |
  >{\columncolor[HTML]{FFFFFF}}c |
  >{\columncolor[HTML]{FFFFFF}}c }
  \hline
                  & FID$\downarrow$   & Top-1$\uparrow$  & Top-5$\uparrow$ & Parameters \\ 
                  &                   & Acc(\%)          & Acc(\%) & (M) \\  \hline
  Ground Truth    & 0     & 76.88 & 93.69 & - \\ \hline
  Grayscale Image & 15.98 & 65.84 & 87.93 & - \\ \hline
  Zhang \cite{DeepExamplar}    & 6.12  & 69.05 & 89.47 & 59.74\\ \hline
  Lu \cite{gray2colornet}      & 6.48  & 66.68 & 87.93 & 53.06\\ \hline
  Blanch \cite{xcnet}           & 6.16  & 68.03 & 89.21 & 29.74\\ \hline
  Yin \cite{transcolor}   & 5.49  & 66.93 & 88.57 & 72.83\\ \hline
  Zhao \cite{Color2Embed}     & 4.30  & 69.57 & 90.18 & 176.52\\ \hline
  Bai \cite{semantic}     & 4.68  & 69.05 & 89.89 & 53.67\\ \hline
  Huang \cite{unicolor}     & 3.79  & 69.88 & 89.74 & 250.5\\ \hline
  SPColor (ours)          & {\bf 3.73}  & {\bf 72.25} & {\bf 91.25} & 146.28\\ \hline
  \end{tabular}
  \end{table}



\section{Experiment}
In this section, we compare our methods with the state-of-the-art approaches both quantitatively and qualitatively. Then, we investigate the effectiveness of the proposed SPC and SMP loss. Besides,  quantitative and quantitative analyses are implemented for the influence of different class number $k$. Moreover, we investigate the robustness of the method and its potential ability in video colorization. Finally, we discuss the limitations of our method, and where future improvements can be made.

\begin{figure*}[!t]
  \centering
  
  \includegraphics[width=\linewidth]{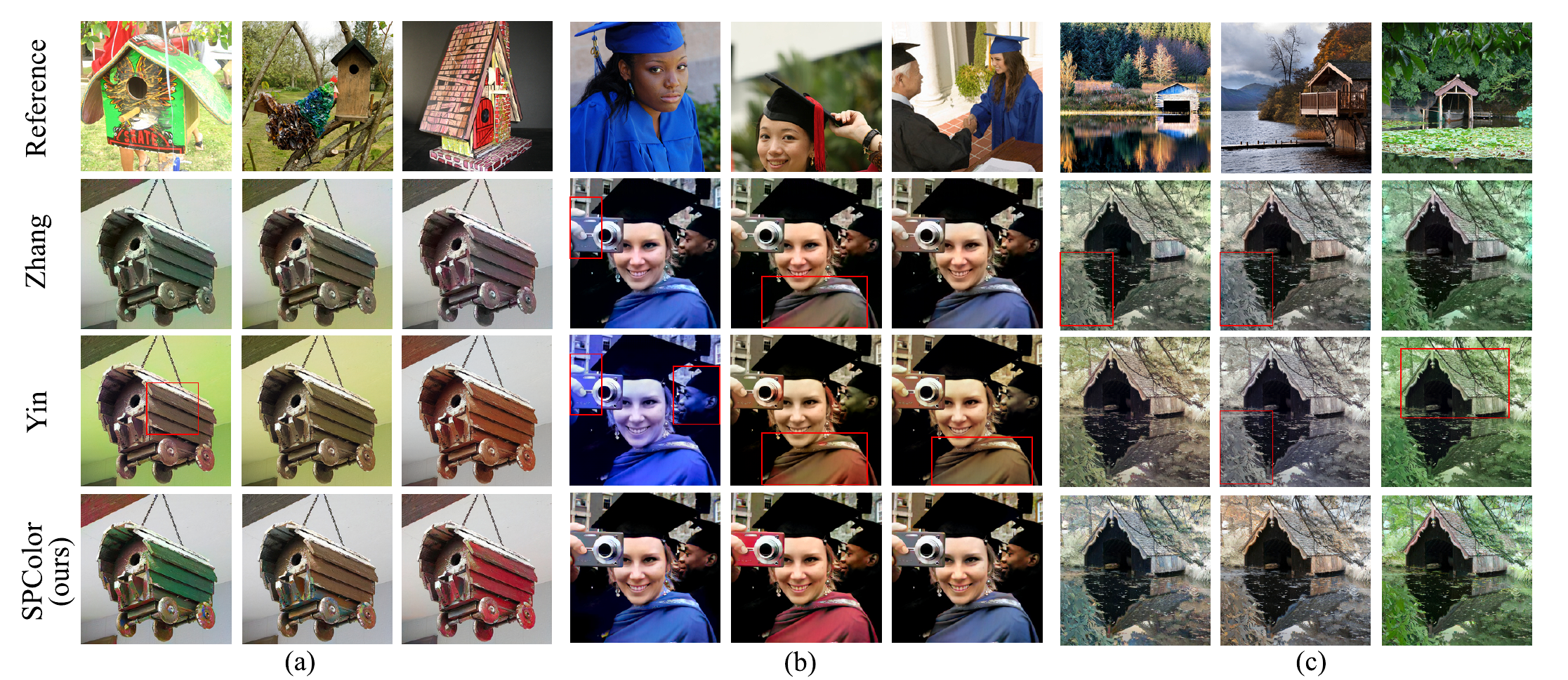}
  
  \caption{Robustness on diverse reference images. We compare SPColor with the methods of Zhang et al. \cite{DeepExamplar} and Yin et al. \cite{transcolor}. Firstly, in (a), Zhang's method is difficult to generate nests of diverse color, and Yin's method also fail on the first reference image that it colorizes the wall to green rather than the nest, while our method gives better results. Secondly, in (b), the other two methods fail to obtain the reference images' color in some place, e.g. the clothes, while our results are more colorful and more similar to the references. Finally, (c) gives several landscape images, the results of the other two methods tend to be monotonous, while our results are more natural and visually appealing.}
  \label{mulref}
\end{figure*}

\begin{figure}[t!]
  \centering
  \includegraphics[width=\linewidth]{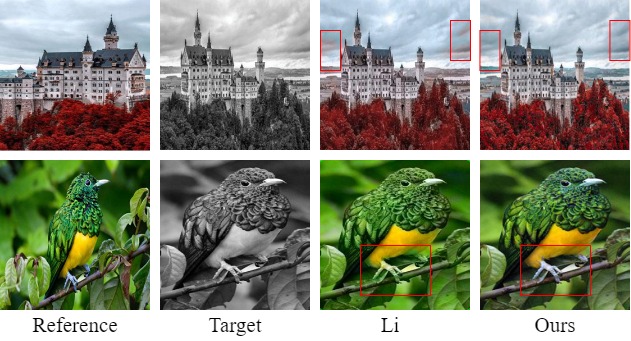} 
  
  \caption{Visual comparison with the method of Li et al. \cite{updown}. As illustrated, \cite{updown} still suffers mismatch even though with the vertical color distribution statistic, while our method produce plausible results.}
  \label{compare_autoexemp}
\end{figure}

\subsection{Comparisons with state-of-the-arts}
We compare our approach against various exemplar-based image colorization methods both quantitatively and qualitatively. The baseline include \cite{DeepExamplar,gray2colornet,xcnet,transcolor,Color2Embed,semantic,unicolor}, which are regarded as state-of-the-art. We use the officially published codes of these methods for comparison, and for \cite{DeepExamplar}, the image colorization mode is adopted. 

{\bf Quantitative comparison}. 
Following the setting of \cite{DeepExamplarimage,DeepExamplar,xcnet}, the performances are evaluated on 3 metrics: FID (Fr´echet Inception Distance) \cite{fid}, Top-1 and Top-5 classification accuracy. The results are reported in {Table \ref{tab:comparision}}. First, FID is utilized to measure the semantic distance between generated image and ground truth. The lower the FID, the more natural the image result. Our method achieves the best FID, which represents that our network generates most perceptually natural results. Second, Top-1 and Top-5 classification accuracy is calculated via a ResNet-101 network \cite{resnet} pretrained on color image classification. They measure how much semantic meanings are contained in generated colors. As shown in {Table \ref{tab:comparision}}, our method obtains the best classification accuracy, indicating that our image result contains the most semantic meaningful colors.
In summary, our approach obtain the best quantitative colorization quality.

{\bf Qualitative comparison}. {Fig. \ref{fig:frontimg}(a)} visualizes some colorization results of each method on ImageNet 10k dataset. More results of qualitative comparison can be seen in Fig \ref{appd:compare}.
The first column and the second column are the reference and target images. And from the third column to the eighth column are the colorization result of Zhang et al. \cite{DeepExamplar}, Lu et al. \cite{gray2colornet}, Blanch et al. \cite{xcnet}, Yin et al. \cite{transcolor}, Zhao et al. \cite{Color2Embed}, Bai et al. \cite{semantic}, and our SPColor respectively.
In the first row, the other methods are easy to colorize the background to brown, while SPColor remains clean background color like reference. In the second row, since the window is not appeared in reference, the other methods tend to colorize it to yellow, while the result of SPColor is obviously more natural. It is the same in the third row. The objects not included in the reference image are easy to have monotonous and unpleasant colors, while SPColor gives better results. Moreover, in the fifth row, the bird which has definite related regions in the reference, is prone to be mismatched to other colors, while by ignoring semantic unrelated regions, SPColor is easy to locate the correct corresponding position. The same situation can be seen in the sixth and the seventh row. In addition, the fourth row shows a combination of the two kinds of mismatch. The sky is prone to be colored by yellow (from the third row to the fifth row) and the bird is easy to be mismatched (in the sixth and seventh row), while our SPColor achieves pleasant result.

We also qualitatively compare the method of Li et al. \cite{updown} with ours. The reference and target images, along with their results, are collected from their published paper. As shown in Fig. \ref{compare_autoexemp}, the results of \cite{updown} still suffer mismatch problem even though with the vertical color distribution statistics from the reference image. This is because that the mismatched colors in these cases actually follow the statistic (the red  under blue in the first row and the green under yellow in the second row), which reflects the limitation of their approach. However, our method is robust to these cases.

To demonstrate the robustness of our method on diverse reference images, we select three different reference images for each target image, and illustrate the colorization results compared with Zhang et al. \cite{DeepExamplar} and Yin et al. \cite{transcolor} in {Fig. \ref{mulref}}. Firstly, in {Fig. \ref{mulref}(a)}, Zhang's method is difficult to generate nests of diverse color, and Yin's method also fail on the first reference image that it colorizes the wall to green rather than the nest, while our method gives better results. Secondly, in {Fig. \ref{mulref}(b)}, the other two methods fail to obtain the reference images' color in some place, e.g. the clothes, while our results are more colorful and more similar to the references. Finally, {Fig. \ref{mulref}(c)} gives several landscape images, the results of the other two methods tend to be monotonous, while our results are more natural and visually appealing.

\begin{table}[]
  \centering
  \caption{Ablation study for our SPC and SMP loss. 
  }
  \label{tab:ablation}
\resizebox{\linewidth}{!}{
  \begin{tabular}{cc|cc|c|c|c}
    \hline
    \multicolumn{2}{c|}{Correspondence}        & \multicolumn{2}{c|}{Loss}                  & FID$\downarrow$  & Top-1$\uparrow$ & Top-5$\uparrow$ \\ \cline{1-4}
    \multicolumn{1}{c|}{Non-local} & SPC & \multicolumn{1}{c|}{Perc. Loss} & SMP Loss &  & Acc(\%) & Acc(\%) \\ \hline
    \multicolumn{1}{c|}{\checkmark} &           & \multicolumn{1}{c|}{\checkmark} &           & 4.98 & 70.75 & 90.13 \\ \hline
    \multicolumn{1}{c|}{}          & \checkmark & \multicolumn{1}{c|}{\checkmark} &           & 4.24 & 70.88 & 91.13 \\ \hline
    \multicolumn{1}{c|}{}          & \checkmark & \multicolumn{1}{c|}{}          & \checkmark & 3.73 & 72.25 & 91.25 \\ \hline
    \end{tabular}
}
  \end{table}

\subsection{Ablation studies}  
To evaluate the effect of the SPC and SMP loss, we replace them with the standard non-local operation and perceptual loss respectively, and the quantitative comparison results are shown in {Table \ref{tab:ablation}}.

\begin{figure}[!t]
  \centering
  
  \includegraphics[width=\linewidth]{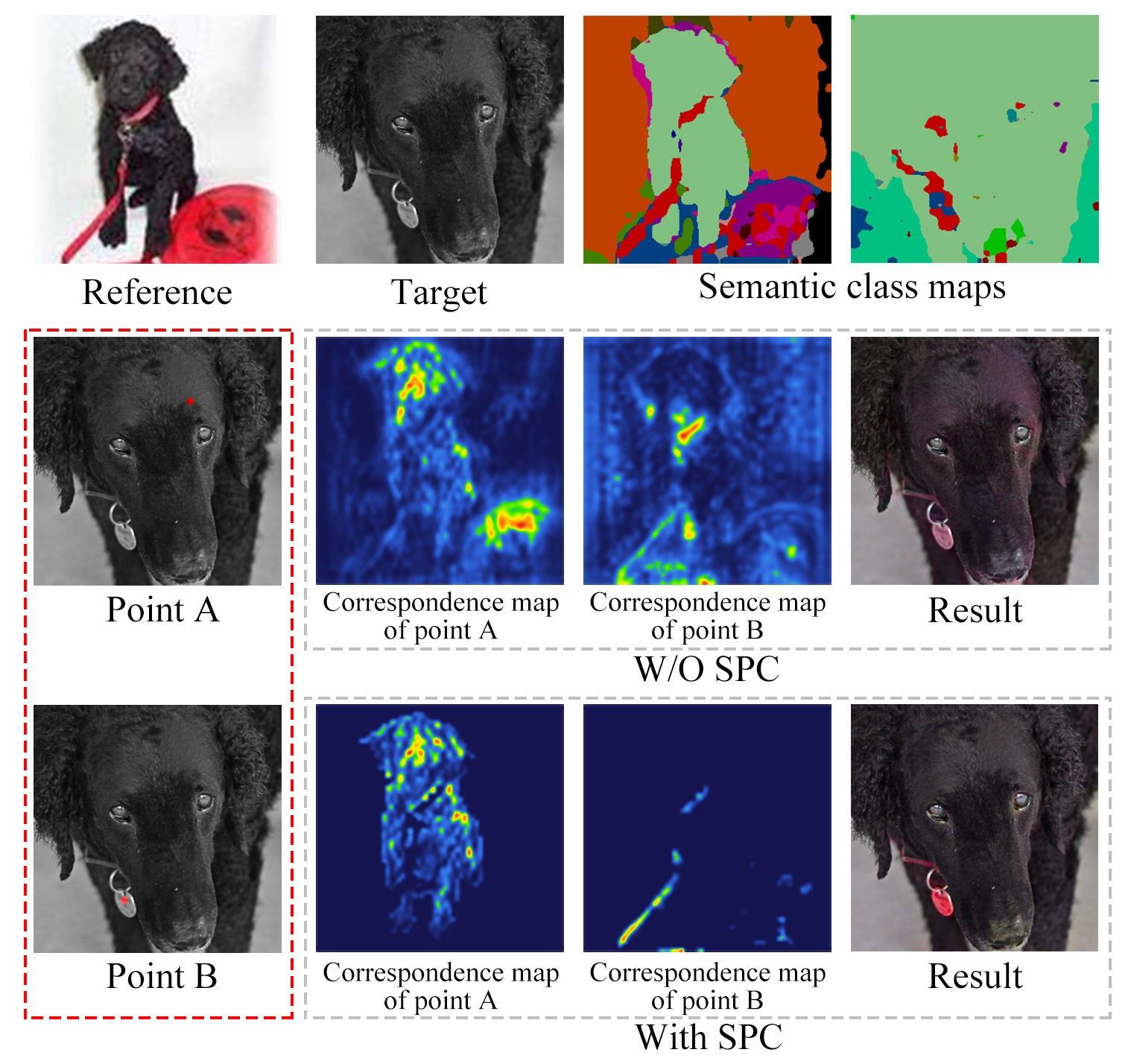} 
  
  \caption{Visualization of the correspondence map. The first line shows the reference, target images and their semantic maps after classification. The second line and the third line plot two points in target image, and show their correspondence maps refer to reference image with or without SPC. 
  }
  \label{cmap}
\end{figure}

{\bf Effect of SPC.} Without semantic prior, SPC degenerates to vanilla non-local operation. Compared with non-local operation, all the three metrics get better with SPC, especially FID and Top-5 accuracy. The obviously improved metrics demonstrate that this operation makes the colorization result more natural and semantically meaningful.
For further analyze the correspondence in SPC, we illustrate the correspondence maps of two points in target image with or without SPC in {Fig. \ref{cmap}}. Without SPC, the point A is likely to correspond to the dog leash, which leads to the red color of point A in result. Besides, the point B's correspondence is easy to be disturbed by the dog or the background, which causes the light color. With SPC, the mismatch is obviously prevented, and the colorization result gets more plausible. 

\begin{figure}[!t]
  \centering
  
  \includegraphics[width=\linewidth]{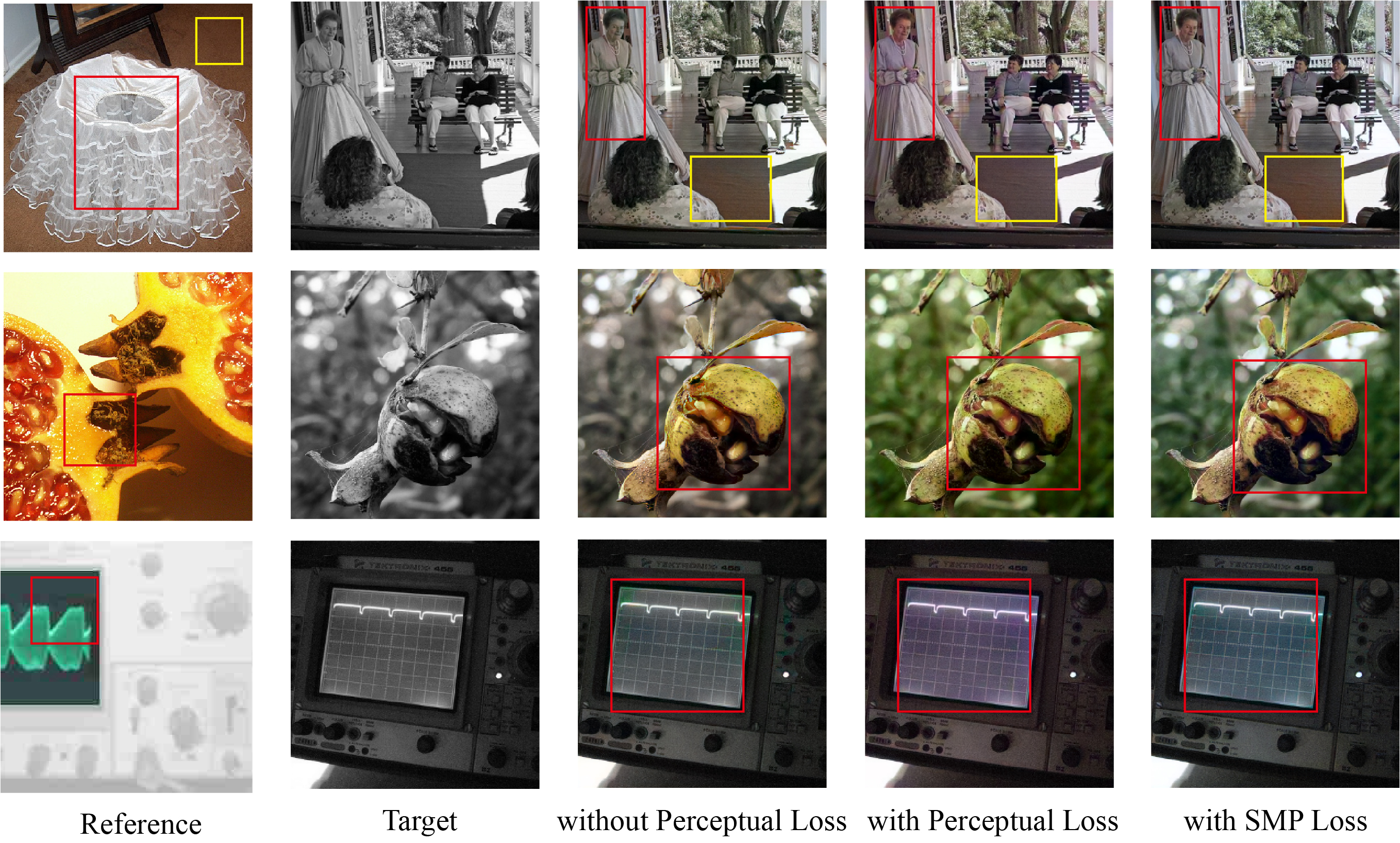} 
  
  \caption{The colorization results with or without perceptual loss and SMP loss. 
  The perceptual loss is effective in generating colors in unrelated regions, which can be seen from the third and the fourth column. However, the colors in related regions may not be similar to the reference image. By applying SMP loss, not only the colors in unrelated regions are generated, but the related regions can also preserve the colors from the reference better.
  }
  \label{smp}
\end{figure}

{\bf Effect of SMP loss.} To analyze the function of SMP loss, we replace the standard perceptual loss with our SMP loss and all the metrics are improved, especially Top-1 accuracy. Besides, as illustrated in {Fig. \ref{smp}}, though the perceptual loss is effective in generating colors in unrelated regions, it will also punish the well related colors, and the generated colors in related regions may not be similar to the reference image. By giving less punishment on the regions with high correspondence and classification confidence, our color result can preserve more on well related colors from reference, which is usually  more distinct and semantically meaningful.

\subsection{Parameter analysis for CRA}
\begin{figure}[!t]
  \centering
  
  \includegraphics[width=\linewidth]{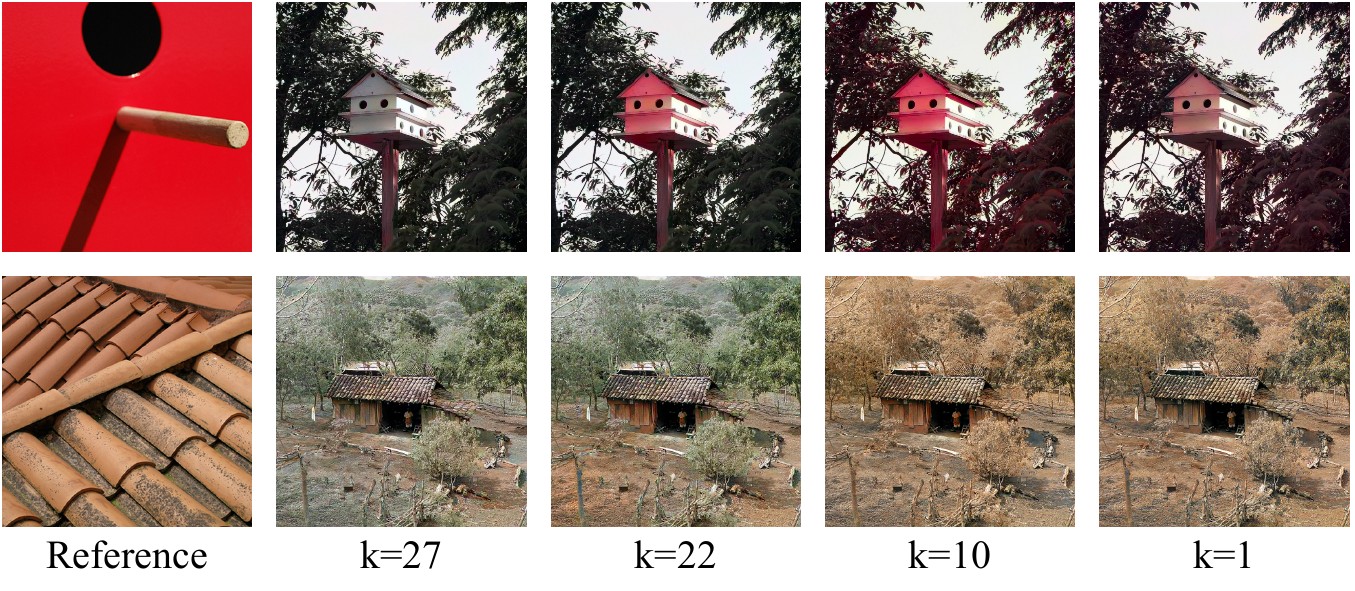} 
  
  \caption{Illustration of colorization with different $k$.}
  \label{fig:cra}
\end{figure}

We change the number of classes $k$ in the CRA and show the qualitatively results on ImageNet 10k in {Table \ref{tab:k}}. Time measures the average processing time of each image in SPC. As $k$ increases, the FID, Top-1 and Top-5 accuracy tend to get better until $k=22$, after which the three metrics goes up and down with small changes, and the processing time continues to grow. This indicates that it is not always better for more classification categories, and the CRA can improves the performance by reduce class number. Moreover, we illustrate some visual results of different $k$ in Fig. \ref{fig:cra}. The selected references contain only a part of the target objects. With the reduction of $k$, more regions from the target images tend to be corresponded with the reference images. For example, in the first row, the nest is more corresponded with the reference in $k=22$ than $k=27$, and both of them obtain plausible results. But in $k=10$ or $k=1$, the tree or the sky are also corresponded, leading to unpleasant colors.  

\begin{figure*}[h]
  \centering
  
  \includegraphics[width=\linewidth]{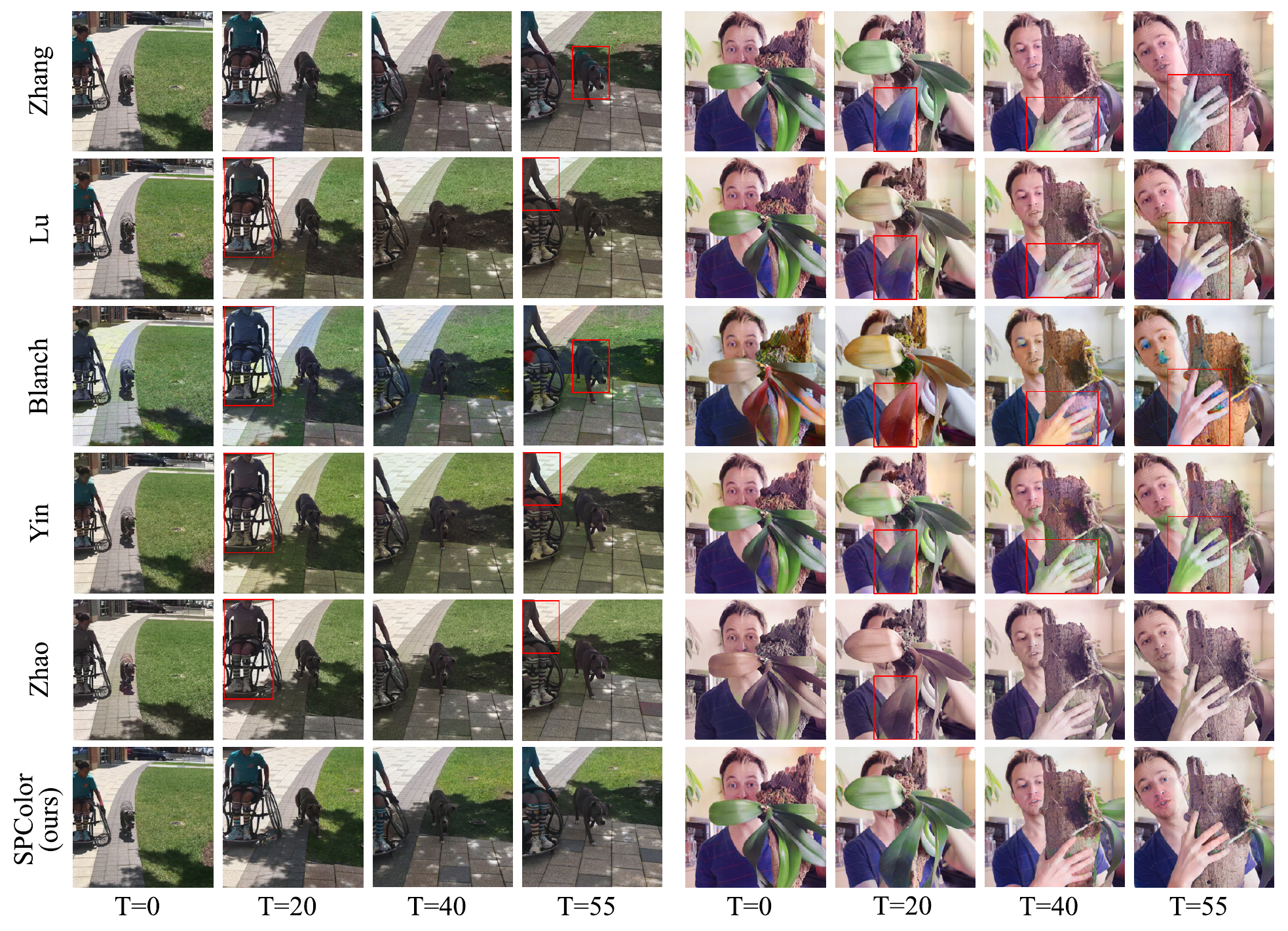}
  
  \caption{Experiment on video color propagation, where the first frame of the video is selected to be the reference image. The video frames are from DAVIS \cite{davis} dataset. From the first row to the sixth \textbf{row} are the colorization result of Zhang et al. \cite{DeepExamplar}, Lu et al. \cite{gray2colornet}, Blanch et al. \cite{xcnet}, Yin et al. \cite{transcolor}, Zhao et al. \cite{Color2Embed} and our SPColor respectively.}
  \label{appd:video}
\end{figure*}

\begin{table}[]
  \centering
  \caption{Quantitative comparison with different $k$. 
  }
  \label{tab:k}
  \begin{tabular}{c|c|c|c|c}
  \hline
  $k$ &
    FID$\downarrow$ &
    \begin{tabular}[c]{@{}c@{}}Top-1$\uparrow$\\ Acc(\%)\end{tabular} &
    \begin{tabular}[c]{@{}c@{}}Top-5$\uparrow$\\ Acc(\%)\end{tabular} &
    \begin{tabular}[c]{@{}c@{}}Time$\downarrow$\\ (ms)\end{tabular} \\ \hline
  1  & 4.336          & 71.13          & 90.47          & 4.5 \\ \hline
  4  & 4.209          & 71.12          & 90.66          & 5.2 \\ \hline
  7  & 4.008          & 71.50          & 91.03          & 5.9 \\ \hline
  10 & 3.882          & 71.71          & 91.09          & 6.2 \\ \hline
  15 & 3.816          & 71.81          & 91.25          & 6.4 \\ \hline
  20 & 3.743          & 72.07          & 91.21          & 6.7 \\ \hline
  21 & 3.748          & 71.91          & 91.21          & 6.7 \\ \hline
  22 & \textbf{3.728} & \textbf{72.25} & 91.25          & 7.1 \\ \hline
  23 & 3.734          & 72.08          & 91.22          & 6.9 \\ \hline
  24 & 3.737          & 72.08          & 91.23          & 6.7 \\ \hline
  25 & \textbf{3.728} & 72.08          & \textbf{91.26} & 7.2 \\ \hline
  26 & 3.732          & 72.19          & 91.25          & 7.2 \\ \hline
  27 & 3.737          & 72.10          & 91.25          & 7.5 \\ \hline
  \end{tabular}
\end{table}

\begin{figure}[h]
  \centering
  
  \includegraphics[width=\linewidth]{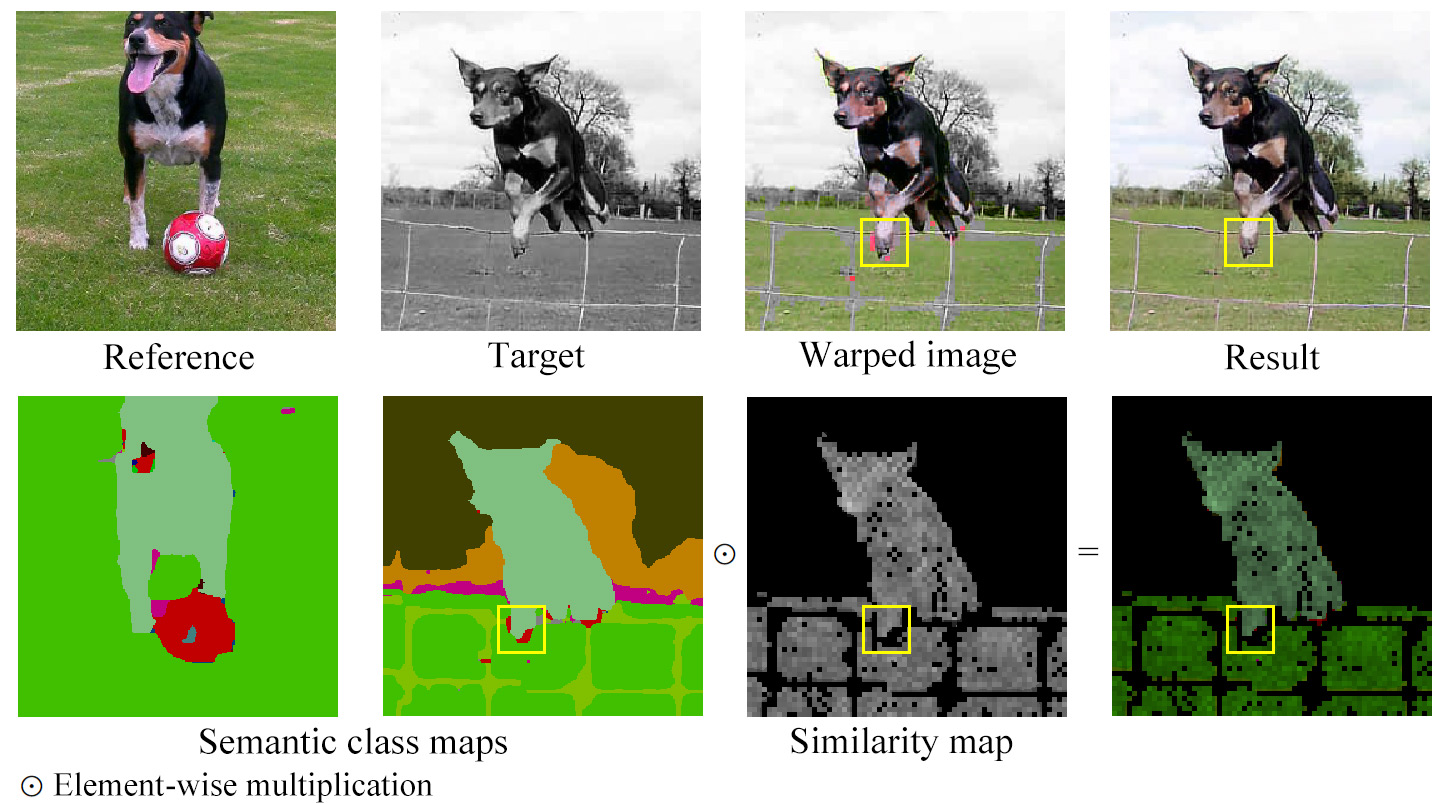}
  
  \caption{Example of the robustness to semantic prior error. The first line contains the reference and target images, the warped image from SPC and the result from colorization network. The images in the second line are the semantic maps, the similarity map from SPC, and its product with target's semantic map. As illustrate, the misclassified pixels have low similarity values since the low classification and corresponding confidence, and the improper colors are fixed in colorization network.}
  \label{appd:error}
\end{figure}

\begin{table}[!t]
  \centering
  \caption{Quantitative results with supervised or unsupervised semantic segmentation model. 
  }
  \label{tab:supervise}
  \begin{tabular}{c|c|c|c|c|c}
  \hline
  Models & Acc.$\uparrow$ & mIoU$\uparrow$ &
    FID$\downarrow$ &
    \begin{tabular}[c]{@{}c@{}}Top-1$\uparrow$\\ Acc(\%)\end{tabular} &
    \begin{tabular}[c]{@{}c@{}}Top-5$\uparrow$\\ Acc(\%)\end{tabular}
     \\ \hline
  Supervised \cite{stego} & 73.75 & 38.85  & 3.78          & 71.83          & 91.27         \\ \hline
  Unsupervised \cite{stego} & 54.41 & 26.32 & 3.74          & 72.09          & 91.24          \\ \hline
  \end{tabular}
\end{table}

\subsection{Robustness to semantic prior errors}
The SPColor uses unsupervised semantic segmentation model in the network for fair comparison with other methods. Utilizing the unsupervised model, it is normal to have few classification errors in processing. However, a stronger supervised semantic segmentation model is also feasible to the approach. {\bf Table \ref{tab:supervise}} shows the classification accuracy and mIoU (Mean Intersection over Union) of the supervised or unsupervised model on the CocoStuff \cite{cocostuff} dataset, and its corresponding colorization performance. For the supervised model, we use the linear-probe mode of the model in \cite{stego}. As illustrated, the supervised model has obviously better classification accuracy. However, the colorization performance of the two model is close. It represents that the very accurate segmentation prior is not such important in our approach, and our model is robust  to the semantic prior errors to some extent.

{Fig. \ref{appd:error}} shows an example of how our method handles these classification errors. As a classification mistake, the pixels approach to the dog legs in target image are classified to the same class as the ball in reference. This lead to improper red colors around dog legs in warped image. However, one may observe from the similarity map and its product with the semantic class map, that the correspondences of misclassified pixels have low similarity values (the pixels in red label disappear after multiplied by the similarity map). This is because that the classification confidence is usually lower in misclassified pixels, and the correspondence are established between two very different objects. Thus, the improper colors are easy to fixed after colorization network, which can be seen from the result image in the first line.

\subsection{Generalization to video colorization}
We further implement our method on video color propagation (where the first frame of the video is selected to be the reference image, and whose colors are then propagated to the following target grayscale frames) to explore the applications of our method on video colorization. The colorized frames are shown in {Fig. \ref{appd:video}}. In this video, the girl's T-shirt is mismatched with the dog in the first and the third column, which colorize the dog to unpleasant blue color. And in the second, the fourth and the fifth column, the color on the girl's T-shirt is faded. Besides, in the right video, with the rotation of the plant, the colors of the leaves are changed in the other methods. And the man's hand, which is not appeared in reference, is prone to be colorized unnaturally. In both of the two video our method gives better results.

\subsection{Limitations}
Though our method obviously enhances the robustness for non-ideal reference image in image colorization, it still has few distinct limitations. Firstly, as seen in {Fig. \ref{spcolor}}, the SPC in our method divided the extracted features to several feature vectors according to the semantic class labels. However, the feature vector are in different length since the number of pixels in a class is indeterminate, which lead to the difficulty for parallel processing, and extended the processing time. Future works can focus on the feature alignment for this issue. Secondly, since a semantic segmentation network is introduced, our model has more parameters than most of recent methods (see in Table \ref{tab:comparision}), combining the semantic segmentation network with the feature extraction network may reduce the amount of parameters. 
Finally, we focus on the problems of mismatch in semantic correspondence in this paper, and exclude the situations that the reference is totally unrelated to the target or even be a color palette. In previous works \cite{DeepExamplarimage,gray2colornet,transcolor}, the problem is resolved by a color database or the reference image's color distribution. There is no conflict between our method and their solutions. Indeed, our method can be applied in their solutions to further improve the performance.

\section{Conclusion}
In this paper, we focus on the mismatch problem in exemplar-based image colorization task. Specifically, novel components are introduced, including a semantic prior guided correspondence network, a category reduction algorithm, and a similarity masked perceptual loss. Furthermore, we propose a category manipulation process to manipulate the constraints of semantic prior, which increases the flexibility of SPC and produces varied results. 
Moreover, in the experiments, we visualize and analyze the effects of the proposed components, and show the effectiveness of our method both quantitatively and qualitatively. Finally, we conclude several limitations of our method, and discuss the future works.

\section*{Acknowledgments}
We would like to thank CCTV Qiyun (Beijing) Technology Co., Ltd. Some real old images of this work are provided by CCTV Qiyun (Beijing) Technology Co., Ltd. This paper also thanks the support of ”the Fundamental Research Funds for the Central Universities” (2022RC53)

\bibliographystyle{IEEEtran}
\bibliography{reference}

\newpage
\appendix
\begin{figure*}[]
  \centering
  
  \includegraphics[width=\linewidth]{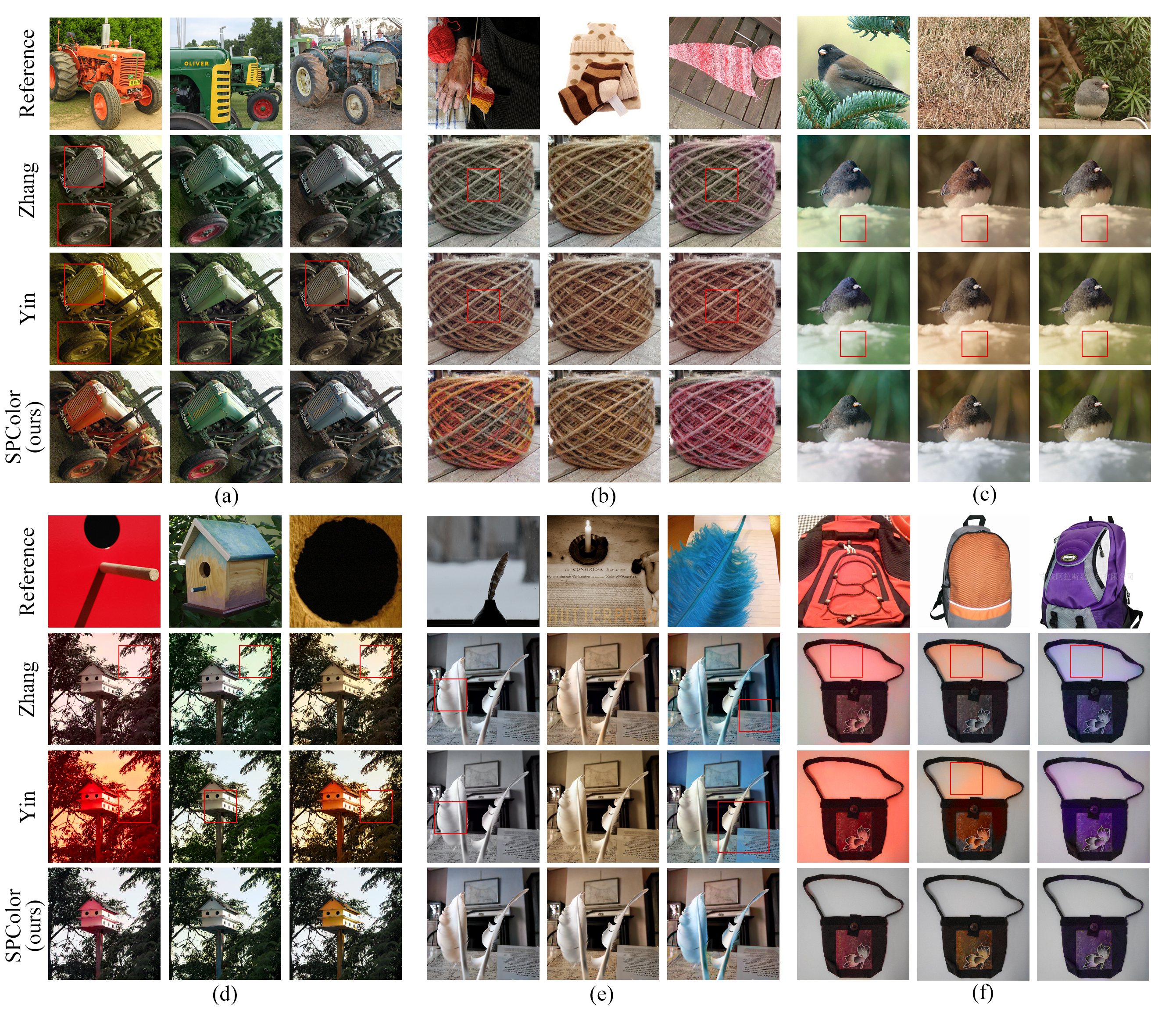}
  
  \caption{More results of colorization on diverse reference images. We compare SPColor with the methods of Zhang et al. \cite{DeepExamplar} and Yin et al. \cite{transcolor}.}
  \label{appd:diffref}
\end{figure*}

\subsection*{Qualitative comparison}
Fig \ref{appd:diffref} shows more results of colorization on diverse reference images. Firstly, in { Fig. \ref{appd:diffref}(a)}, the other two methods fail to obtain the reference images' color in some place, e.g. the tires, while our results are more colorful and more similar to the references. It is the same in { Fig. \ref{appd:diffref}(b)}, the wool are more similar to the references in our results. Moreover, in { Fig. \ref{appd:diffref}(c)}, we show that the color of unrelated regions in our method can be plausible with diverse reference images, while the recent methods tend to generate monotonous colors. It is the same in Fig. \ref{appd:diffref}(d). Furthermore, in Fig. \ref{appd:diffref}(e)(f), the recent methods are easy to cause color bleeding, while SPColor obtains clear results with semantic prior.

\end{document}